\documentclass[sigconf]{acmart}

\AtBeginDocument{%
  }

\usepackage[noabbrev,capitalise]{cleveref}
\usepackage{subfigure}
\usepackage{algorithm}
\usepackage{algorithmic}
\setcopyright{acmlicensed}
\copyrightyear{2025}
\acmYear{2025}
\acmDOI{XXXXXXX.XXXXXXX}
\acmConference[GENNEXT@SIGIR'25]{Make sure to enter the correct
  conference title from your rights confirmation email}{July 17,
  2025}{Padova, Italy}
\acmISBN{978-1-4503-XXXX-X/2018/06}

\newcommand{\todo}[1]{\textcolor{red}{TODO: #1}}

\begin{document}

\title{Asking Clarifying Questions for Preference Elicitation With \\ Large Language Models}

\author{Ali Montazeralghaem}
\email{alimontazer@google.com}
\orcid{1234-5678-9012}
\affiliation{%
  \institution{Google}
  \city{Mountain View}
  \state{CA}
  \country{USA}
}

\author{Guy Tennenholtz}
\email{guytenn@google.com}
\affiliation{%
  \institution{Google}
  \city{Mountain View}
  \state{CA}
  \country{USA}
}

\author{Craig Boutilier}
\email{cboutilier@google.com}
\affiliation{%
  \institution{Google}
  \city{Mountain View}
  \state{CA}
  \country{USA}
}

\author{Ofer Meshi}
\email{meshi@google.com}
\affiliation{%
  \institution{Google}
  \city{Mountain View}
  \state{CA}
  \country{USA}
}

\renewcommand{\shortauthors}{Montazeralghaem et al.}

\begin{abstract}
 Large Language Models (LLMs) have made it possible for recommendation systems to interact with users in open-ended conversational interfaces. In order to personalize LLM responses, it is crucial to elicit user preferences, especially when there is limited user history. One way to get more information is to present clarifying questions to the user. However, generating effective sequential clarifying questions across various domains remains a challenge.
To address this, we introduce a novel approach for training LLMs to ask sequential questions that reveal user preferences. Our method follows a two-stage process inspired by diffusion models. Starting from a user profile, the forward process generates clarifying questions to obtain answers and then removes those answers step by step, serving as a way to add ``noise'' to the user profile. The reverse process involves training a model to ``denoise'' the user profile by learning to ask effective clarifying questions.
Our results show that our method significantly improves the LLM's proficiency in asking funnel questions and eliciting user preferences effectively.
\end{abstract}



\iffalse
\begin{teaserfigure}
  \includegraphics[width=\textwidth]{sampleteaser}
  \caption{Seattle Mariners at Spring Training, 2010.}
  \Description{Enjoying the baseball game from the third-base
  seats. Ichiro Suzuki preparing to bat.}
  \label{fig:teaser}
\end{teaserfigure}
\fi


\maketitle

\section{Introduction}
Recommendation systems (RSs) have become essential tools for making vast amounts of online content accessible to users.
Such systems typically leverage past interactions to learn about a user’s preferences and improve their future recommendations.
Nevertheless, in many cases, information about such preferences is lacking; for example, with new users who have little interaction history, or when privacy constraints limit the use of past interactions. Uncertainty about current user preferences can also result from idiosyncratic factors related to the current context (mood, social setting, etc.).
Rather than relying solely on passive observation of user behavior, an RS can employ \emph{preference elicitation} (PE) techniques by directly asking the user questions which may clarify their preferences \citep{keeney-raiffa,salo2001preference,AlMamunur2008,rao-daume-2018,martin2024model-free}.
PE increases user agency by allowing users to directly communicate their current needs and preferences to the RS, thereby improving the quality of their recommendations.

With the rapid improvement and growing adoption of Large Language Models (LLMs), it is now possible to augment RSs with conversational interfaces, giving rise to Conversational Recommendation Systems (CRS) \citep{friedman2023,he2023,wu2024, montazeralghaem2021large, montazeralghaem2022learning, montazeralghaem2022extracting}.
An important capability of CRS is to perform PE by presenting elicitation questions to users within a multi-turn dialogue. Through direct PE questions, the system can clarify user needs and yield high-quality personalized recommendations.
Simple prompting techniques can direct the LLM to ask elicitation questions whenever appropriate, but is this the best we can do?
In this paper, we study how to optimize LLMs to ask high-quality elicitation questions.

\begin{figure*}[t]
    
  \centering
    \includegraphics[width=1\textwidth]{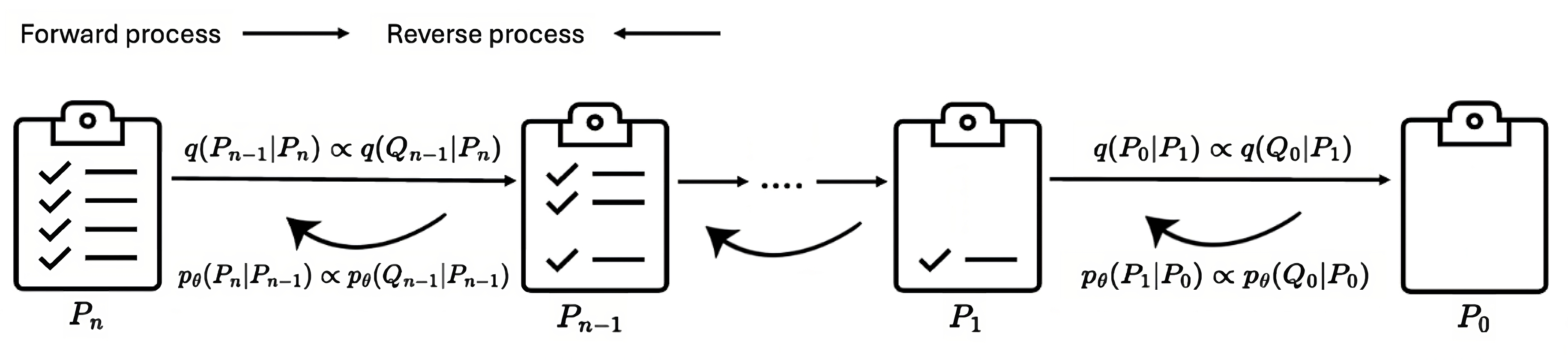}
    \caption{Our model for addressing corrupted user profiles and reconstruction through clarifying questions.}
    \label{fig:gen_profile}
\end{figure*}

Specifically, we propose a framework that starts from a user profile which includes relevant information about the user in text format. We process the user profile in two phases: a forward process and a reverse process, inspired by diffusion models. Specifically, in the forward process we begin by putting the profile information in structured JSON format and ordering the information from most specific to most general. We then generate an elicitation question corresponding to each piece of information in the profile and incrementally remove information from the profile until we are left with an empty user profile. This process is inspired by discrete diffusion models \citep{austin2021structured,reid2023diffuser}, where the input information is corrupted in a forward process and then a model is trained to iteratively denoise the intermediate data until a clean output is obtained in a reverse process.
%For each piece of information removed, we also append an appropriate elicitation question -- the answer to this question corresponds to the information being removed.
We then fine-tune an LLM to ask elicitation questions in order to reconstruct the complete user profile, analogous to the reverse process. Our framework is illustrated in \cref{fig:gen_profile}.

Our experiments demonstrate that this approach yields a highly effective LLM for preference elicitation. Using our generated data, we fine-tune the LLM to produce more effective elicitation questions, which in turn lead to improved reconstruction of the true user profile. Furthermore, the model learns to ask funnel questions—starting with general concepts and gradually becoming more specific as the conversation progresses.
%For evaluation, we let the LLM interact with simulated users and we find that for this task it is also important to fine-tune the simulator.

\section{Background}
\label{background}
\subsection{Text Generation}
In the autoregressive paradigm for text generation, the probability of an entire sequence $s = [s_0, s_1, ..., s_N]$ can be modeled as the product of the conditional probabilities of predicting each token in the sequence from left to right. Mathematically, this is expressed as \citep{bengio2000neural}:
\begin{equation}
    P(S) = \prod_{i=0}^{N} p(s_t|s_0, s_1, \ldots, s_{t-1})
\end{equation}
where $p(s_t|s_0, s_1, \ldots, s_{t-1})$ represents the conditional probability of generating the token $s_t$ given all the preceding tokens $s_0, s_1, ..., s_{t-1}$.
\subsection{Profiling Process}
Let \( P = (P_0, P_1, \ldots, P_n) \) represent a series of \( n \) versions of a profile containing user information gathered over time, where \( P_0 \), \( P_i \), and \( P_n \) denote the initial, intermediate (at timestep \( i \)), and final versions of the profile, respectively.
We can model the probability of this series of profile versions occurring consecutively as follows:
\begin{equation}
    p(P) = \prod_{i=0}^{n}p(P_i|P_{0},\ldots,P_{i-1}).
\end{equation}

\subsection{Diffusion Models}
We see a connection between profiling processes and diffusion models \citep{sohl2015deep, ho2020denoising}. In computer vision, diffusion models work by starting with a noisy image and gradually refining it until it becomes clear and complete. Continuous diffusion models are often trained by modeling a Markov chain $x_T, \dots, x_t, \dots, x_0$, where $x_0$ represents the original image and $x_T$ corresponds to pure Gaussian noise. This chain is generated by progressively adding Gaussian noise to $x_t$ to obtain $x_{t+1}$, a process known as the forward or corruption process. A model parameterized by $p_\theta$ is then trained to reverse this process, effectively ``denoising'' $x_{t+1}$ back to $x_t$, thereby reconstructing the original input from noisy representations.

%Analogized to text, this allows us to formulate profiling user preferences as a discrete diffusion process, where a null user profile or a prototype is iteratively refined into a complete user profile. Our model draws inspiration from this process (see \cref{fig:gen_profile}). Similarly, in our approach, we start with ``noisy'' (empty) user profiles and improve them step by step by asking relevant questions, ultimately creating a clearer picture of the user’s preferences.
We formulate user preference profiling as analogous to a discrete diffusion process, often used in text generation. In such processes, a basic or null state is iteratively refined into a complete output. Our model adopts this concept (see \cref{fig:gen_profile}): we start with "noisy" (initially empty) user profiles and refine them incrementally by asking relevant questions, leading to a more precise understanding of user preferences.

\section{Proposed Model}
%\todooinAli{Ofer and Guy, I used this paper as a reference to structure paper and write the proposed model section: \url{https://openreview.net/pdf?id=nG9RF9z1yy3}, please let me know if there is a problem.}
\label{our_model}

Our goal is to optimize an LLM to ask good clarifying questions.
Taking inspiration from diffusion models in discrete spaces \citep{reid2023diffuser}, we start from a complete (textual) profile and gradually remove information from the profile in a \emph{forward process} until it is empty, akin to corrupting the profile. We use an LLM to generate an elicitation question that would reveal the information being removed from the profile in each forward step. We then apply a \emph{reverse process} that maps a partial profile at time step $t$ to the elicitation question corresponding to the next piece of information from the profile. We hypothesize that an LLM trained using the reverse process can ask good clarifying questions based on partial information about the user.
The order of the elicitation questions plays an important role in the flow of the conversation. Importantly, we want to ask more general questions (e.g., ``what is your favorite genre?") before asking more specific questions (e.g., ``do you like artist X?"). To this end we use an LLM to order the information in the user profile from least general to most general in the forward pass, so that in the reverse pass we get the desired \emph{funneling} effect.

For evaluation, we train another LLM to simulate the user's responses to the generated elicitation questions. The user model is given the full ground-truth user profile and generates replies to elicitation questions according to the profile (it also leaves a question unanswered if the profile does not contain information related to the question). We use this model to evaluate our approach by generating sessions where the PE model (trained in the reverse process) interacts with the simulator, and compare the resulting profile to the ground-truth profile.

In the next subsection, we next formulate the problem, starting with the Reverse process. Then, in the Forward process, we explain how to %corrupt user profiles and 
generate training data for the Questioner and user simulator.

\subsection{Profile Reconstruction by Asking Questions}
Our generative process is trained using the Sequential Question Answering (SQN) process. Particularly, in SQN, the objective is to find appropriate questions to transform an initial empty user profile \( P_0 = \emptyset \) into a final ground-truth profile \( P_n \) through a sequence of intermediate profiles \( P_1, \ldots, P_{n-1} \).
Each profile \(P_t\) represents the state after \(t\) question-answer interactions and is defined as the set of collected pairs accumulated up to that point: \( P_t = \{(Q_i, A_i)\}_{i=0}^{t-1} \).
That is, given a potentially corrupted or partial profile \( P_t \) (where \(t < n\)), our goal is to learn the generative process by which the complete profile \(P_n\) is formed.

Using the chain rule and assuming conditional independence, we have:
\begin{equation}
    p_{\theta, \phi}(P_n) = \prod_{t=1}^{n}p(P_{t}|P_{t-1};\theta, \phi)
\end{equation}

The probability \( p(P_{t} \mid P_{t-1}; \theta, \phi) \) is expressed as a function of three components:

\begin{equation}
\begin{aligned}
    p(P_{t} \mid P_{t-1}; \theta, \phi) &= p_\theta(Q_{t-1} \mid P_{t-1}) \\
    &\quad \times p_\phi(A_{t-1} \mid Q_{t-1}, P_{t-1}) \\
    &\quad \times p(P_{t} \mid P_{t-1}, Q_{t-1}, A_{t-1})
\end{aligned}
\end{equation}

where:
\begin{itemize}
    \item \( p_\theta(Q_{t-1} \mid P_{t-1}) \) is the probability of generating the question \( Q_{t-1} \) given the partial profile \( P_{t-1} \), parameterized by \(\theta\). This reflects the process of the Questioner.
    \item \( p_\phi(A_{t-1} \mid Q_{t-1}, P_{t-1}) \) is the probability of providing the answer \( A_{t-1} \) to the question \( Q_{t-1} \), given the question \( Q_{t-1} \) and the partial profile \( P_{t-1} \), parameterized by \(\phi\). This reflects the answerer's response (i.e., user simulator).
    \item \( p(P_{t} \mid P_{t-1}, Q_{t-1}, A_{t-1}) \) is the probability of generating the next state \( P_{t} \) given the previous state \( P_{t-1} \), the question \( Q_{t-1} \), and the answer \( A_{t-1} \). This component is deterministic and does not have any learnable parameters, it can be expressed as:
\begin{equation}
\resizebox{0.45\textwidth}{!}{
  $
  \begin{aligned}
  {\scriptstyle p(P_{t} \mid P_{t-1}, Q_{t-1}, A_{t-1})} & {\scriptstyle =}
  \begin{cases} 
  {\scriptstyle 1} & {\scriptstyle \text{if } P_t = P_{t-1} \cup \{(Q_{t-1}, A_{t-1})\}} \\
  {\scriptstyle 0} & {\scriptstyle \text{otherwise}}
  \end{cases}
  \end{aligned}
  $
}
\label{eq:transition_prob}
\end{equation}
    This probability is 1 if the new profile \( P_t \) is obtained by adding the question \( Q_{t-1} \) and answer \( A_{t-1} \) to the previous profile \( P_{t-1} \). Otherwise, it is 0.
    %\todo{[Ofer] Should we define that $P_t=\{(Q_i,A_i)\}_{i=1}^{t-1}$ earlier when we introduce $P_t$?}

\end{itemize}

Note that updating the profile by adding answers alone is possible, but our experiments show that including questions along with answers in user profiles helps the model avoid repetitive queries and improves its performance. It also allows using simple yes/no answers.

Our objective is to maximize the probability of generating the complete user profile by asking effective clarifying questions which can be formalized as follows:

\begin{equation}
%(\hat{\theta}, \hat{\phi}) = \arg
\max_{\theta, \phi} \sum_{i=1}^{|I|} \log(p_{\theta, \phi}(P_n^i))~,
\end{equation}
where $P^i_n$ is the complete profile of user $i$, and $|I|$ is the number of users. We optimize this objective by fine-tuning two LLMs: one as a questioner and another as a user simulator to answer questions generated by the Questioner.
Note that we could have used a pretrained user simulator, however we have found that fine-tuning the simulator significantly improves the results. % (see \cref{sec:experiments}).

\subsection{Profile Corruption}
\label{sec:profile_corruption}
In the forward process, we are given a text representation of the information \( P^u \) of user \( u \) (e.g., preferred movie genres or movies previously watched and their descriptions). Our goal is to gradually add noise until the profile contains no information. %to this information until the profile is destroyed.
Common operations for adding noise in such discrete spaces include inserting, deleting, and replacing words \citep{reid2023diffuser}. In this work, we focus on deletion. We propose to ask a clarifying question at each step based on the partial user profile and then \emph{remove} the answered portion from the user profile.

Given information in text format \( P^u \) from user \( u \), we first convert this information into a structured format (e.g., JSON) to create a structural user profile.
This conversion allows us to use structured tags and values, enabling the efficient querying and manipulation of profile data and simplifying operations in subsequent processes.
We can generate these structured formats of user profiles using an LLM: $JP^{u} := \text{LLM}(P^u)$.

Given \( JP^{u} \), we now aim to generate questions in the forward process and create partial profiles. The generation of questions can vary, but we adhere to the following two constraints:

\begin{enumerate}
    \item In the reverse process, questions are asked starting from the easier and more straightforward ones, progressing to more specific questions.
    \item We consider dependencies between questions. If there is a dependency, we first ask about broader aspects before more specific ones. For example, in movie recommendations, we first inquire about the movie genre before asking about more specific preferences, such as sub-genres or favorite directors.
\end{enumerate}

We examine different approaches for generating questions from user profiles. We found that ranking the tags in the JSON user profile based on the notion of generality and then prompting the LLM to generate \textit{funnel Questions}, which start from more general questions and gradually proceed to more fine-grained ones, yields the best results in terms of satisfying the constraints.

Specifically, let $JP^u=\{(t_i,c_i)\}_{i=1}^m$ be the structured profile, with tags $t_1,\ldots,t_m$ and the corresponding information content $c_1,\ldots,c_m$. For example, in the movie domain, we can have $t_i=\texttt{'Genre'}$, $c_i=\texttt{'The user likes action movies'}$. We first use an LLM to rank the tags from general to specific. We then use the tag ranking to generate questions using a specific prompt\iffalse(see Appendix~\ref{sec:appendix_prompt} for details)\fi:
$
    (Q_0, A_0), \ldots, (Q_{n-1}, A_{n-1}) = \text{LLM}(JP^{u}, \{t_1, t_2, t_3, \ldots, t_m\}),
$
where $Q_i$ denotes the generated question, and $A_i$ denotes its corresponding answer derived from the user profile (e.g., $Q_i=$\texttt{'Do you like action movies?'}, $A_i=\texttt{'yes'}$). Crucially, we define a mapping $\mathcal{T}(Q_i,A_i)$ that identifies the set of tag-content pairs from the original profile $JP^u$ which are directly addressed or answered by question $Q_i$ and answer $A_i$. Formally, $\mathcal{T}(Q_i,A_i) = \{(t_k, c_k) \in JP^u \mid (t_k, c_k) \text{ is addressed by } Q_i, A_i \}$. For example, for $Q_i$=\texttt{'Do you like action movies?'} and $A_i=\texttt{'yes'}$, $\mathcal{T}(Q_i,A_i)$ might correspond to $\{($\texttt{'Genre'}, \texttt{'The user likes action movies'}$)\}$ or potentially multiple related pairs if the question covers more ground. To simplify the notation, we sometimes denote \(\mathcal{T}_i=\mathcal{T}(Q_i,A_i)\).
%\todo{[Ofer] this is a bit confusing. Perhaps we can define a mapping from $(Q_i,A_i)$ to a set of tag/content pairs instead of letting $A_i$ be this mapping. This will make \cref{eq:partial_profile} easier/clearer.}

Since we generate questions in a funnel-like manner, the questions \( Q_i \) are ordered such that they start from broad concepts (e.g., \( Q_0 \)) and progress to more detailed questions (e.g., \( Q_{n-1} \)). Here, \( n \) represents the total number of questions generated for the user profile to gather as much information as possible. Note that the number of tags \( m \) and the number of questions \( n \) can differ. The LLM may generate multiple questions about a single tag or generate questions that address multiple tags simultaneously.

Based on the generated questions, we need to create data through the forward process and then utilize it in the reverse process.
Since the questions are generated in a funnel manner, question \( Q_0 \) is more general compared to question \( Q_{n-1} \) in the forward process. Therefore, in the forward process, we begin with \( Q_{n-1} \), assuming the user profile is complete. Subsequently, we remove the information corresponding to the answer to this question \( Q_{n-1} \) from the user profile. This process continues, removing the information related to each question's answer until the user profile is empty.

Formally, the partial user profile at step can be represented as follows:
\begin{equation}
    \label{eq:partial_profile}
    JP^u_{t} = JP^{u} \setminus \bigcup_{i=t}^{n-1} \mathcal{T}_i
\end{equation}
where \( t \) ranges from \( n \) (representing the full profile, as the union is empty) down to 0 (representing the empty profile, as information from all questions $Q_0..Q_{n-1}$ has been removed). Here, \( JP^u_{t} \) denotes the partial user profile available just before asking question $Q_t$ (or at the end if $t=n$), and \( JP^{u} = JP^u_n \) is the complete initial user profile. Note that $JP^u_0 = \emptyset$.

So, by using the partial user profiles \(JP^u_t\), the data that we can use in the reverse process for user \( u \) to train the model would be:
\begin{equation}
    D_u = \{(Q_{n-1}, JP^u_{n-1}), (Q_{n-2}, JP^u_{n-2}), \ldots, (Q_{0}, JP^u_{0})\}.
\end{equation}
Here, each pair \((Q_i, JP^u_t)\) represents a training instance where, given the partial user profile \( JP^u_t \) as input, the model should generate the corresponding question \( Q_i \) as the target.

Let's consider \( I = \{u_1, u_2, \ldots, u_{|I|}\} \) as the set of all users. After generating Funnel questions for all user profiles, the data for the reverse process is:
\begin{equation}
    D = \{D_{u_1}, D_{u_2}, \ldots, D_{u_{|I|}}\}.
\end{equation}
We use this data for fine-tuning the Questioner.
The forward process is summarized in \cref{alg:profile_corruption}.

\begin{figure*}[h]
  \centering
    \includegraphics[width=0.96\textwidth]{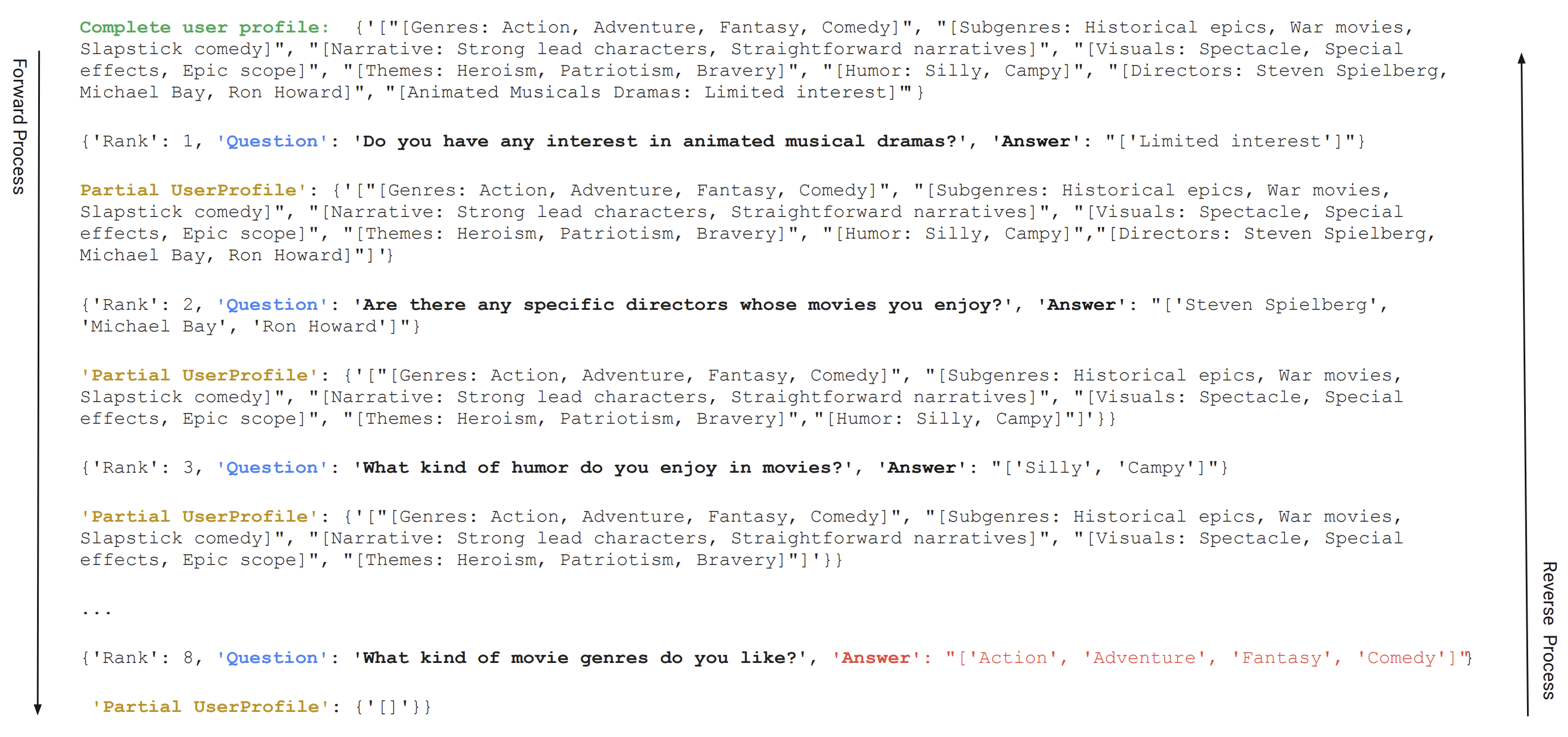}
    \caption{An example of our framework for user profile processing. Starting from a complete user profile in textual form, the forward process converts it into structured JSON and sequentially generates elicitation questions while progressively removing information. The reverse process then reconstructs the profile by iteratively answering the elicitation questions}
    \label{fig:example}
\end{figure*}

\begin{algorithm}[t]
\caption{Forward process: Profile Corruption}
\label{alg:profile_corruption}
  \begin{flushleft}
    \textbf{Input}: A user profile \( P_u \) in text format.\\
    \textbf{Output}: Training data \( D_u \), comprising question-partial user profile pairs for various partial profiles.\\
  \end{flushleft}
\begin{algorithmic}[1] % The number tells where the line numbering should start
    \STATE Convert \( P_u \) into a JSON format \( JP^{u} \).
    \STATE Sort tags \( \{t_1, t_2, \ldots, t_m\} \) from \( JP^{u} \) based on notion of generality.
    \STATE Generate Funnel Questions \( \{(Q_0, A_0), (Q_1, A_1), \ldots, (Q_{n-1}, A_{n-1})\} \) based on the extracted tags.
    \STATE $t \gets n - 1$
    \WHILE{$t \geq 0$}
        \STATE Create partial profile \( JP^{u_t} \) by using \cref{eq:partial_profile}.
        \STATE \( D_u \gets D_u \cup \{(Q_t, JP^{u_t})\} \) %Add the pair \( (Q_t, JP^{u_t}) \) to the training data \( D_u \).
        \STATE \( t \gets t - 1 \) %Decrease \( t \) by 1
    \ENDWHILE
    \STATE \RETURN \( D_u \)
\end{algorithmic}
\end{algorithm}

\subsubsection{User Simulation}
%To evaluate our approach, we need an environment to interact with our Questioner and answer questions based on user preferences. This environment requires ``real users'' to chat with the Questioner and provide answers.
Evaluating our approach requires an environment where our `Questioner' model interacts with a `user simulator' that answers questions based on specific user preferences.
Creating such an environment for research is challenging. Even on platforms with millions of users, launching a dialogue system without extensive training for real users may fail \citep{gao2021advances,Jannach2020}.

To address this problem, a common practice is to use LLMs as simulated users due to their ability to answer questions effectively. Therefore, in each conversation between our Questioner and user simulator, we provide the user profile to the LLM and ask it to find the answer from the profile if possible, that is, $A=LLM(P,Q)$. Otherwise, we instruct the LLM to respond with ``I don't know'' in cases where the answers are not present in the user profile, assuming that the user does not have specific preferences regarding the asked question.
%\todooin{Consider: $A=LLM(P,Q)$ for consistency. Plus, include prompt in Appendix and ref here.}

To enhance the ability of the LLM to answer questions more effectively, we fine-tune it using the data generated in
profile corruption %\cref{sec:profile_corruption}
as follows: %(see \cref{sec:experiments}):
\begin{equation}
    \begin{aligned}
        \hat{D}_u = & \{(\mathcal{T}_{n-1}, Q_{n-1}, JP^{u}), (\mathcal{T}_{n-2}, Q_{n-2}, JP^{u}), \\
              & \ldots, (\mathcal{T}_0, Q_{0}, JP^{u})\}
    \end{aligned}
\end{equation}
%\todo{[Ofer] I defined $\hat{D}_u$ to distinguish from $D_u$.}
Each tuple $(\mathcal{T}_i, Q_i, JP^{u})$ serves as a training instance for fine-tuning the user simulator. In other words, given the question $Q_i$ and the user profile $JP^{u}$, the model should find the corresponding answers $A_i$ (chain-of-thought) and mapping $\mathcal{T}_i$ (response output) as the target.
An example illustrating the full process is shown in \cref{fig:example}.

\begin{algorithm}[t]
\caption{Evaluation Process}
\label{alg:eval_process}
  \begin{flushleft}
    \textbf{Input}: A corrupted profile \( P_t \), target profile \( P_n \), parameters \( \theta, \phi \), maximum question number \( T \) \\
    \textbf{Output}: A sequence of questions and answers that transforms \( P_t \) to \( P_n \)
  \end{flushleft}
\begin{algorithmic}[1]
    \STATE Initialize profiles: \( P_{\text{current}} \gets P_t \)
    %\STATE Initialize an empty list for questions \( Q \) and answers \( A \)
    \STATE Initialize question count: \( \text{count} \gets 0 \)

    \WHILE{(\( P_{\text{current}} \neq P_n \)) \textbf{and} (\( \text{count} < T \))}
        \STATE Generate question \( Q_{\text{t-1}} \) using the fine-tuned model as the Questioner
        \STATE Query the user simulator model, which accesses the target profile \( P_n \) to determine an answer \( A_{\text{t-1}} \):
        
        \IF{answer is found in \( P_n \)}
            \STATE Set \( A_{\text{t-1}} \) to the corresponding value in \( P_n \)
        \ELSE
            \STATE Set \( A_{\text{t-1}} \) to ``No Preference"
        \ENDIF
        %\STATE Add \( A_{\text{t-1}} \) to the current profile \( P_{\text{current}} \)

        \STATE Update:
            \STATE \( P_\text{current} \gets P_{\text{current}} \cup \{(Q_{\text{t-1}}, A_{\text{t-1}})\} \)
            %\STATE \( Q \gets Q \cup Q_{\text{t-1}} \)
            %\STATE \( A \gets A \cup A_{\text{t-1}} \)
            %\STATE \( P_{\text{next}} \gets P_{\text{current}} \cup \{(Q_{\text{t-1}}, A_{\text{t-1}})\} \)
            %\STATE Append \( Q_{\text{t-1}} \) to \( Q \)
            %\STATE Append \( A_{\text{t-1}} \) to \( A \)
            %\STATE \( P_{\text{current}} \gets P_{\text{next}} \)
            \STATE Increment question count: \( \text{count} \gets \text{count} + 1 \)
    \ENDWHILE
    
    %\IF{\( \text{count} \geq T \)}
    %    \STATE Log warning: ``Question threshold exceeded"
    %\ENDIF
    
    \STATE \RETURN \( P_\text{current} \) %sequence of questions \( Q \) and answers \( A \) leading to \( P_n \)
\end{algorithmic}
\end{algorithm}

\section{Experiments}
\label{sec:experiments}
%\subsection{Dataset}
We use \textit{Movielens}, a movie recommendation dataset widely used in research related to recommender systems \citep{harper2015movielens}. Throughout our experiments, we use the user profiles from \citet{tennenholtz2024demystifying,jeongfactual,tennenholtz2024embedding}. These ground-truth profiles were generated using an LLM and the complete raw history of ratings from each user. The user-profiles were evaluated to be predictive of user ratings in the dataset (see \citep{tennenholtz2024embedding} for more information).

We use the Gemma LLM (7B version) \citep{team2024gemma} with $28$ layers as our Questioner and user simulator (in the reverse process and for fine-tuning) because its weights are publicly available and it has been shown to be one of the best models for its size.

For generating data in the Forward process, we use Gemini $2.0$ \citep{team2023gemini}, which is a larger LLM with greater capability, to ensure that the generated training data for fine-tuning is of high quality.
For fine-tuning, we use Parameter-Efficient Fine-Tuning (PEFT) in our experiments and apply Low-Rank Adaptation (LoRA) \citep{hu2021lora} in all our experiments.
For training our models, we fix the batch size to $64$ and the learning rate to $0.001$.

\begin{figure*}[h]
    \centering
    \subfigure[]{
        \includegraphics[width=0.4\textwidth]{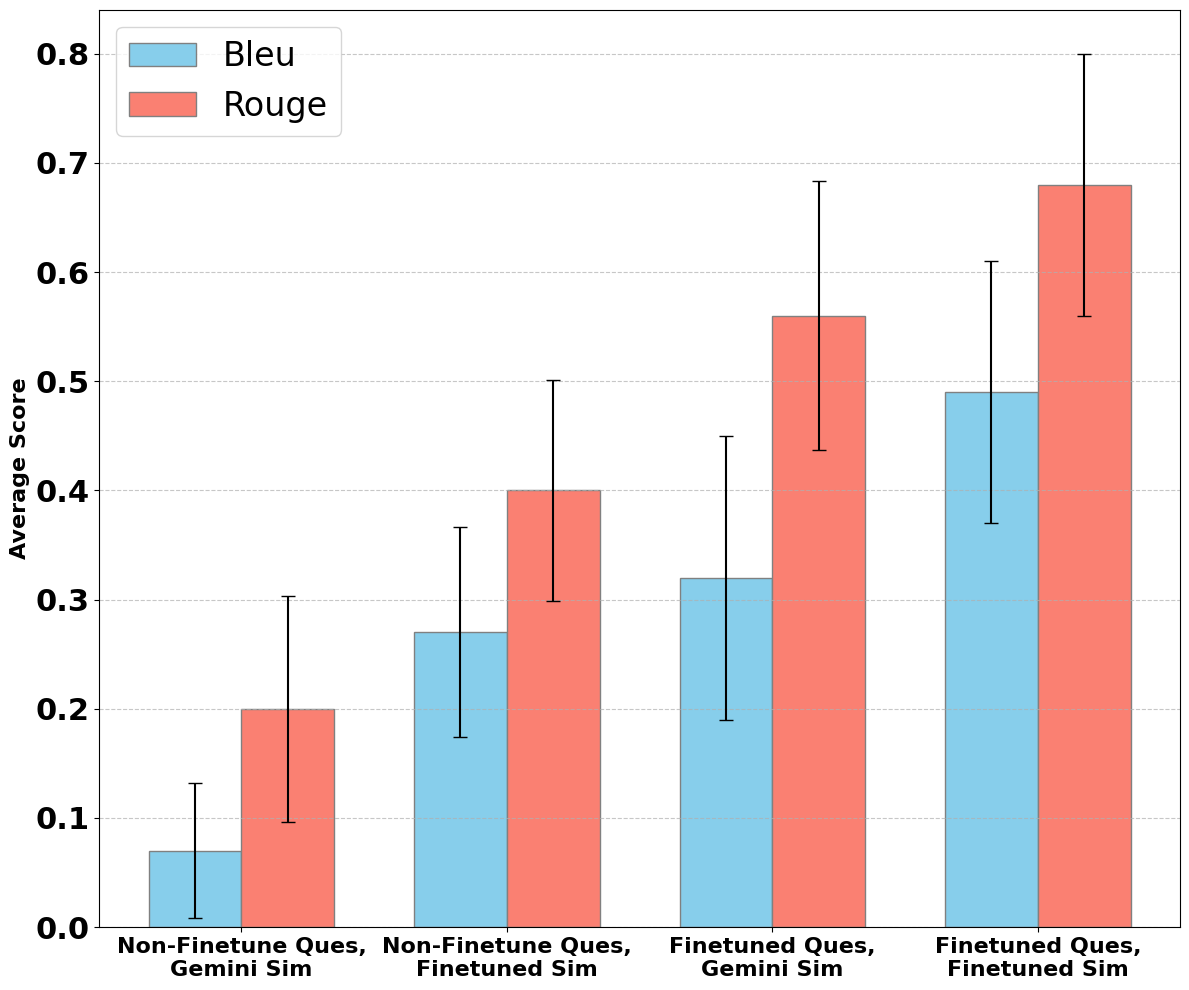}
        \label{fig:first_res}
    }
    \hspace{0.05\textwidth} % Adjust space between figures
    \subfigure[]{
        \includegraphics[width=0.4\textwidth]{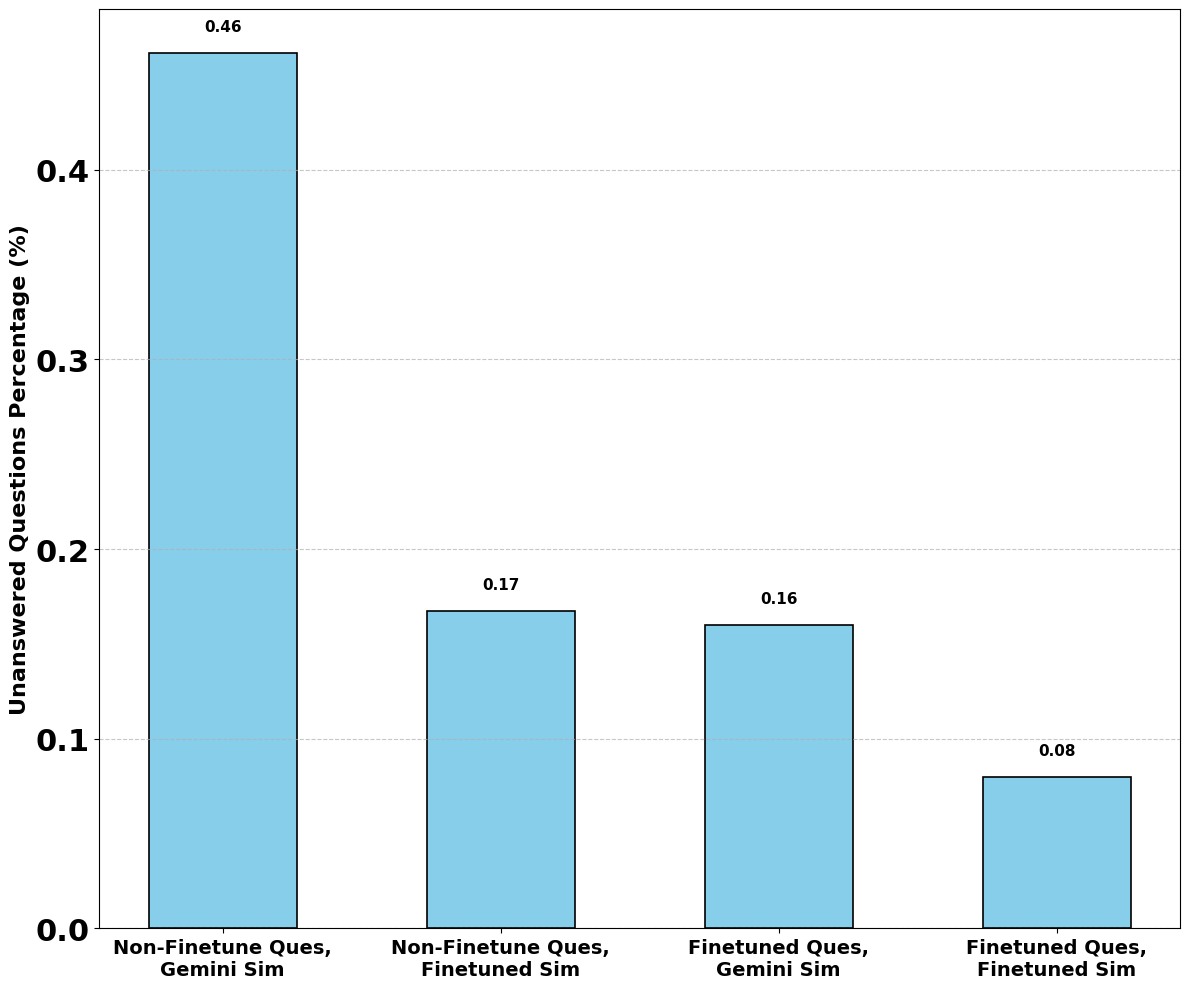}
        \label{fig:unanswered_questions}
    }
    \caption{(a) BLEU and ROUGE scores for four models, showing the performance of the non-fine-tuned and fine-tuned Questioners with different user simulators. (b) Percentage of unanswered questions for models.}
    \vspace{-0.4cm}
    \label{fig:sidebyside}
\end{figure*}

\begin{figure*}[h]
  \centering
    \includegraphics[width=0.9\textwidth]{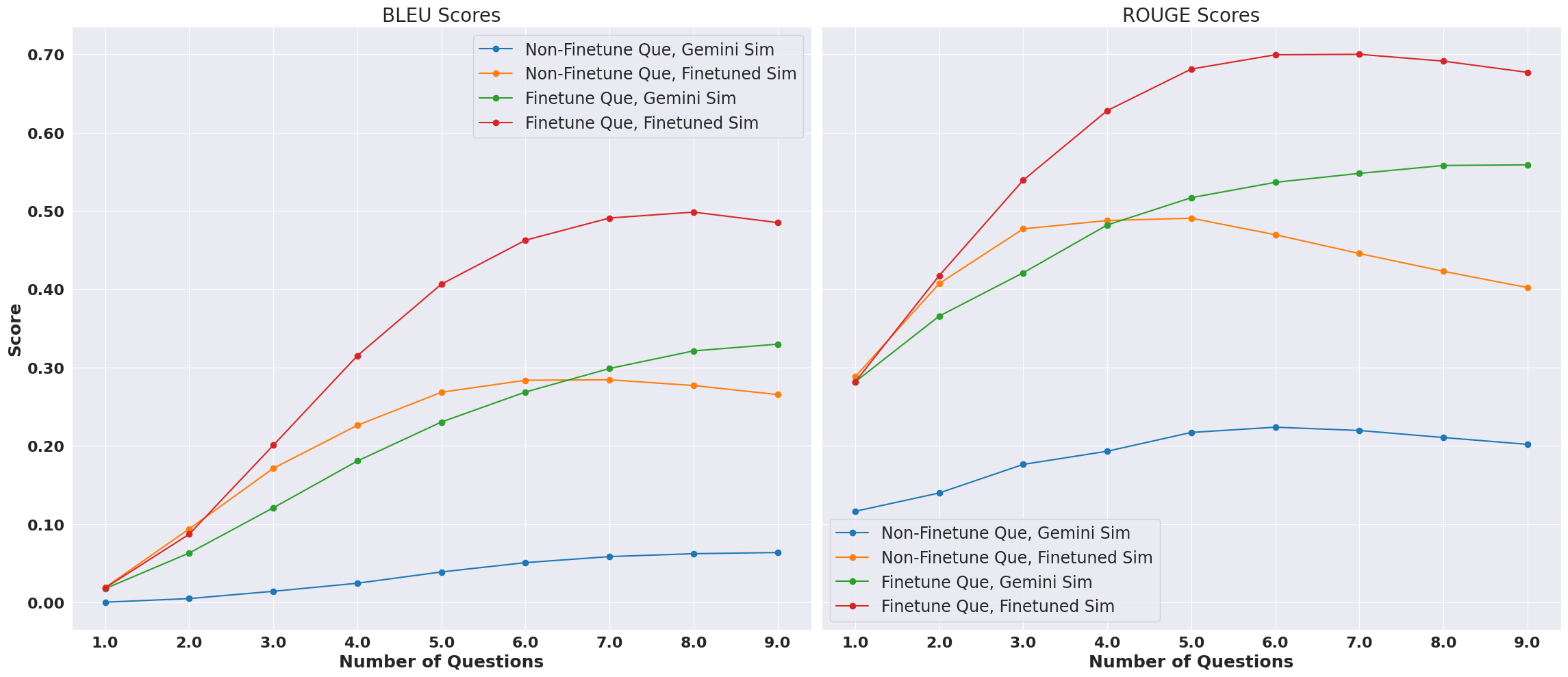}
    \caption{BLEU (left) and ROUGE (right) scores vs. number of questions. \iffalse \todo{[Ofer] 1) The fine-tuned and non-fine-tuned lines seem to be swapped in the legend; 2) The final values don't seem to match the values in \cref{fig:first_res}}\fi}
    \label{fig:effect_questions}
    \vspace{-0.2cm}
\end{figure*}

\subsection{Evaluation Process}
For evaluating the generated questions and their quality, we use our fine-tuned Questioner in an interaction with a user simulator, starting with an empty profile. In each turn of the interaction, the Questioner asks a clarifying question, and the user simulator tries to answer the question given the target (ground-truth) user profile. If the user simulator finds the answer to the question in the user profile, we add it to the current user profile and use it in the next turn; otherwise, we set the answer to the question as ``No Preference.'' This process will continue until the number of questions exceeds a limit (10 questions) or the current user profile matches the target profile. Finally, we compare the generated user profile to the target profile and evaluate it.
To measure the quality of the generated questions, we compare the generated profile to the ground-truth profile using the ROUGE and BLEU metrics.
\Cref{alg:eval_process} shows the evaluation process.
\begin{figure*}[h]
    \centering
    \subfigure[Comparing the overall performance of models by integrating questions and answers (Q-H) into user profiles.]{
        \includegraphics[width=0.4\textwidth]{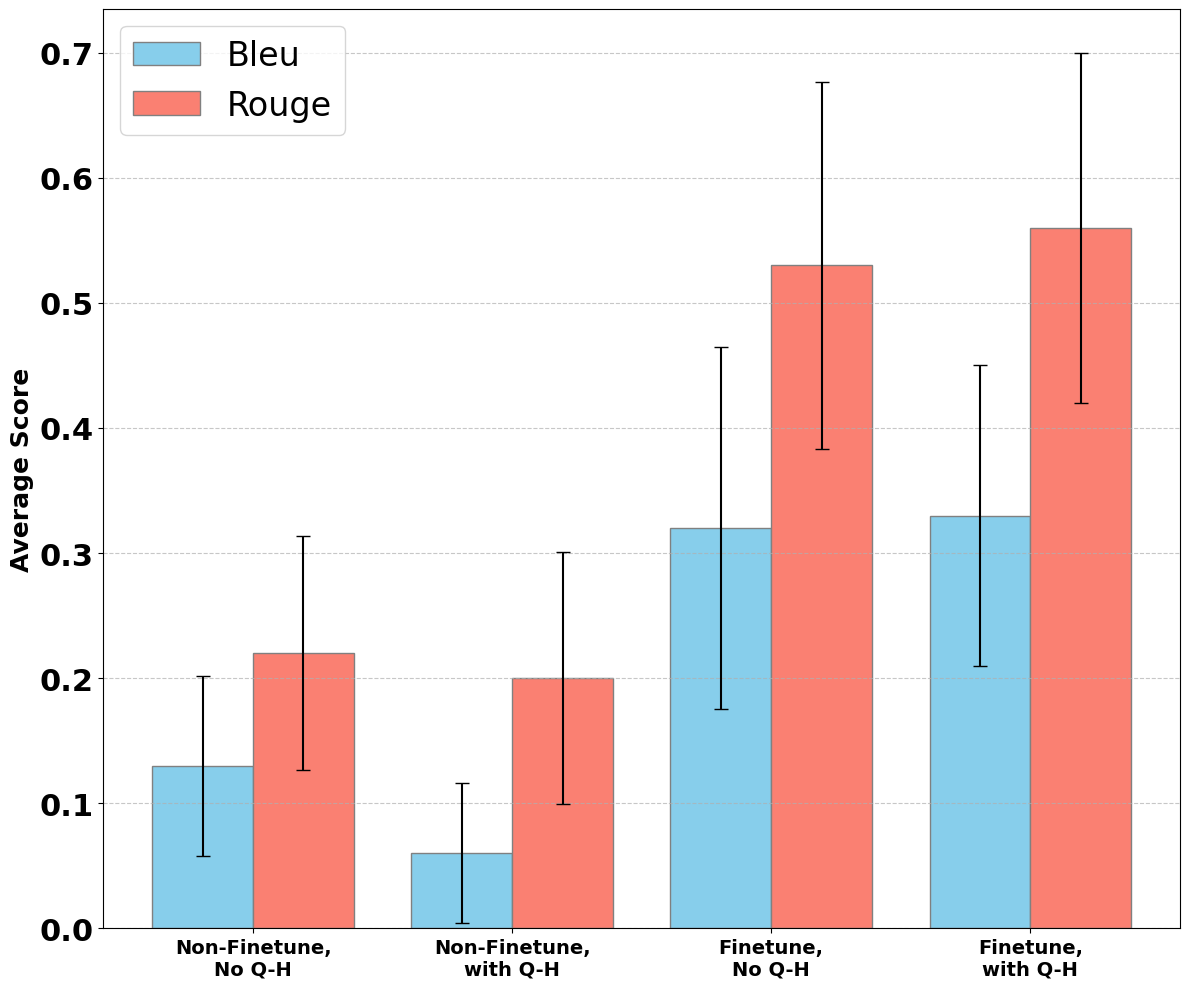}
        \label{fig:added_history}
    }
    \hspace{0.05\textwidth} % Adjust space between figures
    \subfigure[Percentage of repetitive questions]{
        \includegraphics[width=0.4\textwidth]{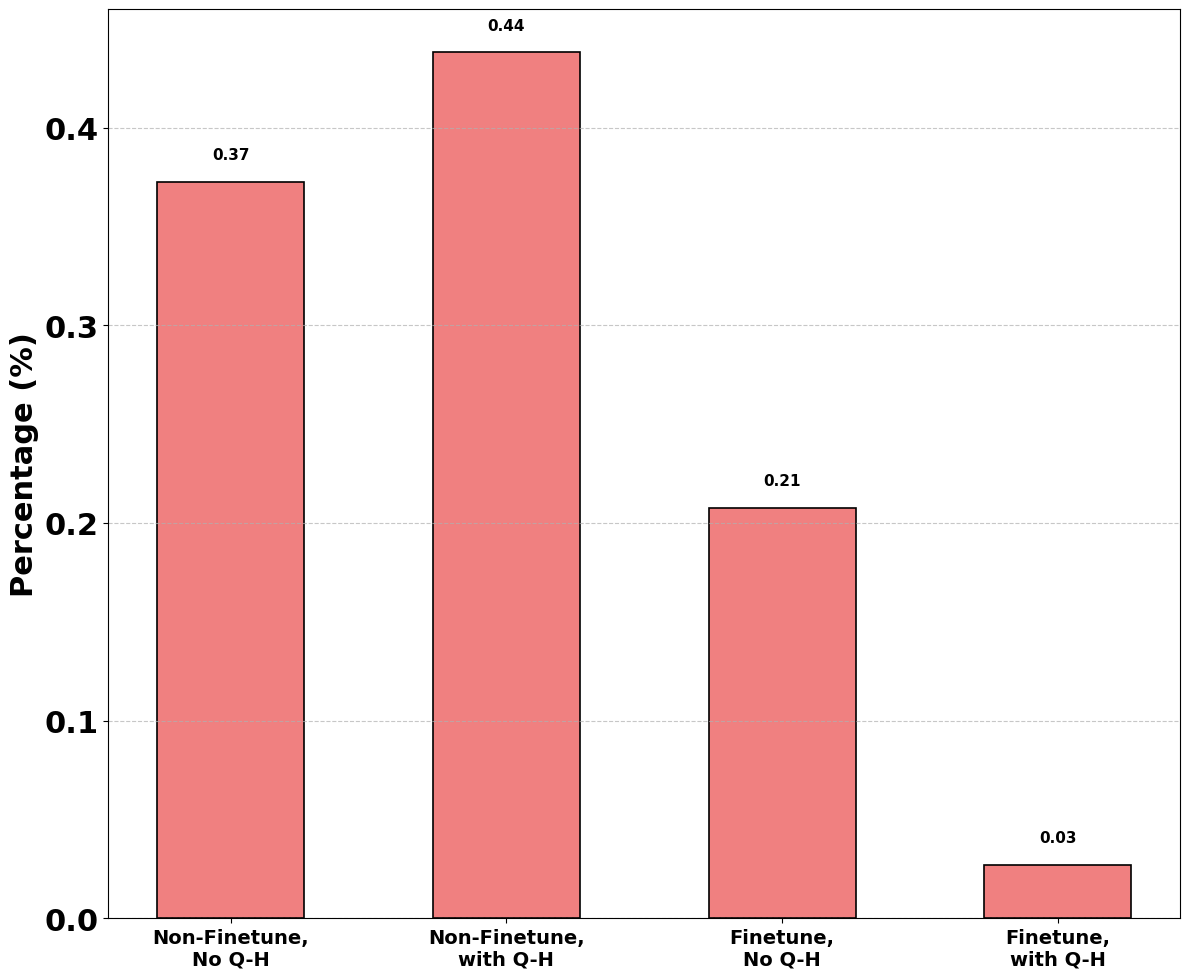}
        \label{fig:added_history_per}
    }
    \caption{Effect of adding questions along with answers to
partial user profiles }
    \label{fig:sidebyside}
\end{figure*}

\begin{figure}[h]
  \centering
    \includegraphics[width=0.4\textwidth]{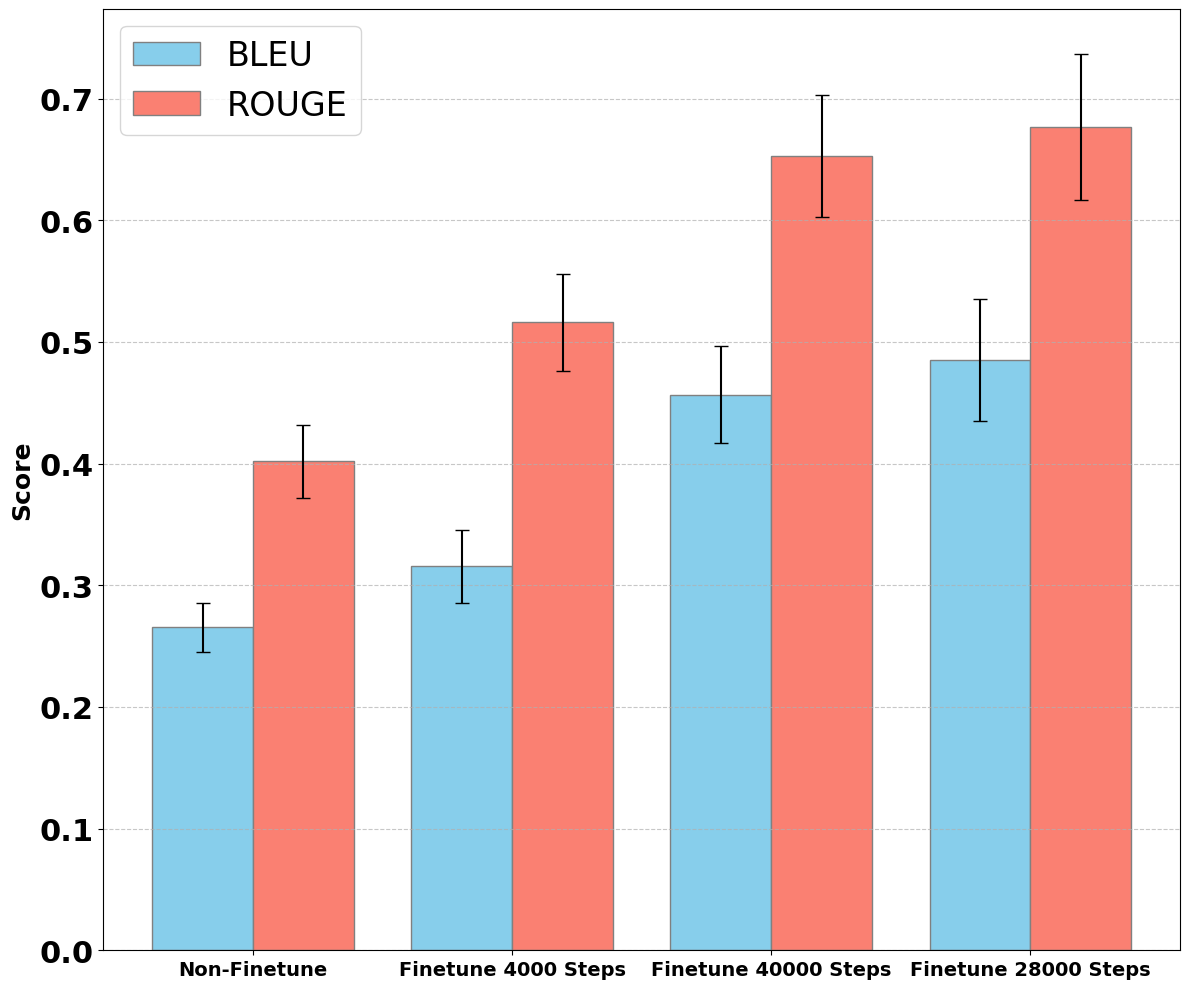}
    \caption{BLEU and ROUGE scores of the Questioner model at different fine-tuning steps (0, 4000, 28000, and 40000)}
    \label{fig:exp_steps}
    \vspace{-0.2cm}
\end{figure}

\begin{figure*}[h]
  \centering
    \includegraphics[width=0.8\textwidth]{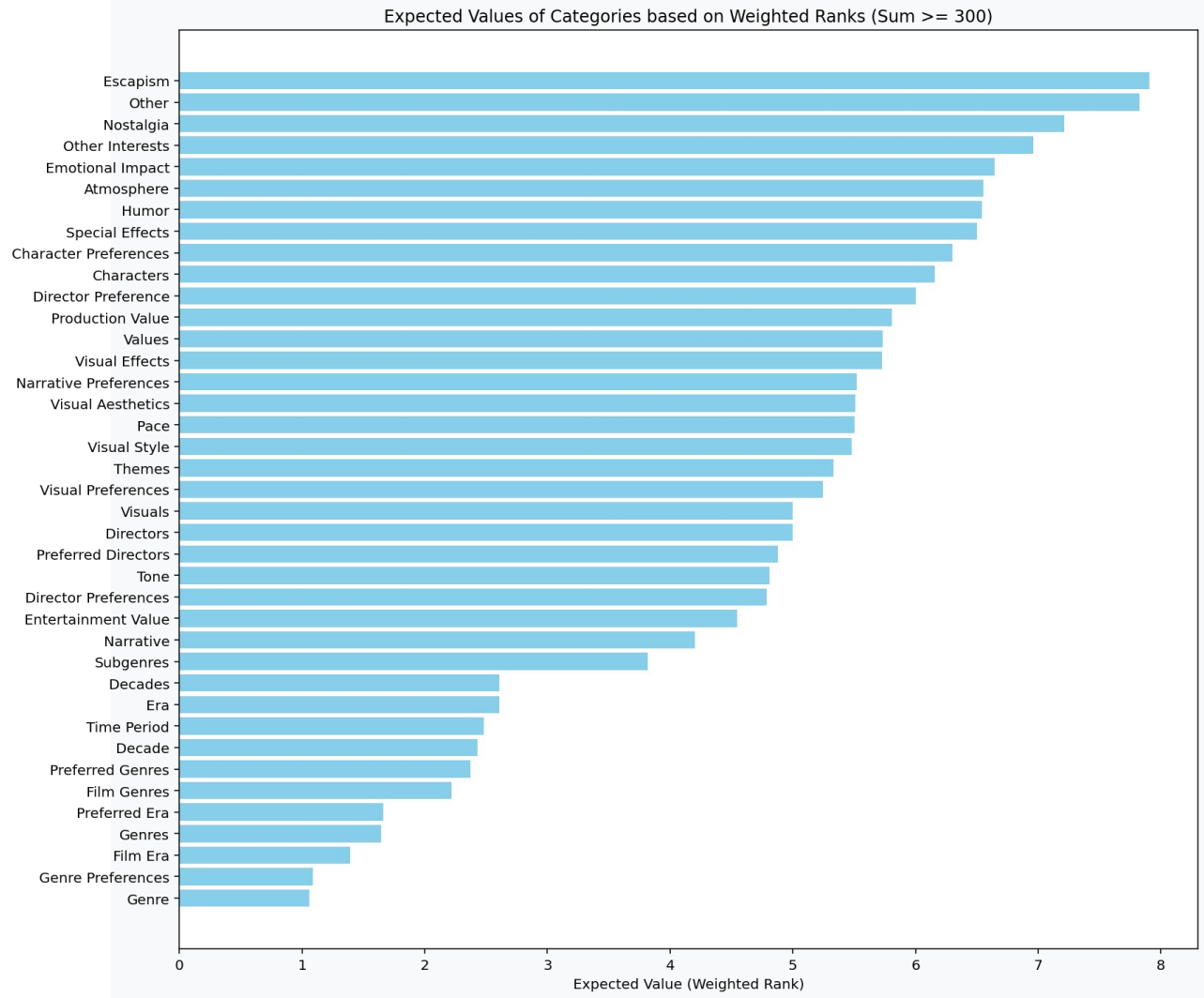}
    \caption{Weighted ranks (WR) of conversation concepts across question positions, illustrating the distribution of each concept within a funnel format. Lower WR values indicate concepts commonly addressed earlier in the conversation, while higher WR values correspond to concepts typically addressed later.}
    \label{fig:exp_funnel_concepts}
\end{figure*}

\subsection{Results and Analysis}

\subsubsection{Effect of fine tuning}
We first compare our fine-tuned Gemma model in the reverse process, using the data generated in the forward process, with a non-fine-tuned Gemma model to evaluate the quality of the generated questions in both models. We also use a non-fine-tuned Gemini model as a user simulator to compare it with our fine-tuned Gemma user simulator. 

The results of this experiment are shown in \cref{fig:first_res}. To observe the effect of the fine-tuning process for the Questioner, we can compare 'non-finetuned questions' and 'finetuned questions' bars in \cref{fig:first_res}. The results show that fine-tuning the model with the generated data can significantly improve its performance (Rouge from $0.4$ to $0.68$ and Bleu from $0.28$ to $0.49$, with finetuned simulation).
This shows that the fine-tuned Questioner is able to ask more effective sequential questions to capture personal information from the user and create the user profile. Similarly, we can observe the same results in the first and third bars, where the simulator is a non-fine-tuned Gemini model instead of a fine-tuned Gemma.

%The Questioners can ask more effective questions, but to better demonstrate the effect of these questions, we could interact the fine-tuned model with real users. However, this would be more expensive and time-consuming. Instead,
Specifically,
we fine-tuned the user simulator to answer the questions more effectively. To compare the results, we can examine the third and fourth bars (or similarly, the first and second bars) in \cref{fig:first_res}. 
The results show that the fine-tuned user simulator can answer questions more effectively compared to the non-fine-tuned Gemini model. Note that for fine-tuning the user simulator, we used Gemma, which is a much smaller model compared to Gemini.

To further analyze these results, we present the percentage of unanswered questions (where the simulator is unable to find the answers in the profile) for four models, as shown in \cref{fig:unanswered_questions}.
This figure shows that the fine-tuned user simulator is able to answer questions from both the non-fine-tuned and fine-tuned Questioners more effectively. In contrast, using the non-fine-tuned Gemini as a user simulator sometimes leads to unanswered questions, which can negatively impact the performance of the Questioner. This highlights the importance of the user simulator interacting with the Questioner.

\subsubsection{Effect of Number of Questions}
To illustrate the effect of the number of questions on BLEU and ROUGE scores across different approaches, we present these scores with varying numbers of questions in Figure \ref{fig:effect_questions}. We compare two types of Questioners: fine-tuned and non-fine-tuned models, as well as two types of simulators: Gemini and fine-tuned simulator. The worst-performing model for asking clarifying questions and answering them is the non-fine-tuned Questioner with the Gemini simulator. This poor performance is due to two factors: the Questioner cannot ask effective questions, and the Gemini simulator is unable to find relevant answers in the user profile. However, by replacing the Gemini simulator with a fine-tuned simulator (represented by the orange line), we observe a performance boost. This indicates that while some of the questions asked by the non-fine-tuned model are effective, their effectiveness is enhanced when answered by a fine-tuned simulator.

By using the Gemini simulator with a fine-tuned Questioner (represented by the green line), we observe a performance boost compared to the non-fine-tuned Questioner with the Gemini simulator. This highlights the effect of fine-tuning the Questioner, enabling it to ask more effective clarifying questions.

The figure demonstrates that our best model (i.e., the fine-tuned Questioner paired with the fine-tuned simulator) excels by asking questions that gather broader information during the initial five turns (up to turn 5), then shifts toward asking more specific and detailed questions in the later turns (turns 6 or 7).
\iffalse
To demonstrate the impact of the number of questions on BLEU and ROUGE scores across different approaches, we present these scores for varying question counts in \cref{fig:effect_questions}. \textcolor{blue}{This figure illustrates that our best model (i.e., fine-tuned Questioner coupled with a fine-tuned user simulator) asks questions that gather a broader range of information in the initial five turns (up to turn 5), followed by more specific and detailed questions in the subsequent turns (6 or 7).} \todo{[Ofer] I think it actually shows that 5 questions are enough and there is little value in asking further.}

From the results, we can also see the effect of fine-tuning the user simulator. In both models (fine-tuned and non-fine-tuned Questioner), using the fine-tuned user simulator can improve performance and demonstrate the importance of having a more effective user simulator.
\fi

\iffalse
\begin{figure}[t]
  \centering
\includegraphics[width=1\textwidth]{figs/main_res_8.png}
    \caption{BLEU and ROUGE scores of the Questioner model at different fine-tuning steps (0, 4000, 28000, and 40000)}
    \label{fig:exp_steps}
\end{figure}
\fi

\subsubsection{Effect of Adding Question History}
In updating the user profile, we either add only the answers to the questions or include both the questions and their answers. In our experiments, we found that adding questions along with answers (see \cref{eq:transition_prob}) can help the model avoid asking repetitive questions.

\cref{fig:sidebyside} shows the results of this experiment. In this experiment, we added questions along with answers in both non-fine-tuned and fine-tuned Questioners. In our experiments, we observe that adding question history can increase performance in a fine-tuned model but decrease it in a non-fine-tuned model. As mentioned above, adding question history to the partial user profile helps the model ask non-repetitive questions. In the fine-tuned model, this can help the model ask non-repetitive and effective questions in the next turns. However, this can reduce the performance of the non-fine-tuned model since it starts to ask non-repetitive questions, but the new questions are not effective because the model is not trained to ask effective questions to reveal user preferences in a sequential manner.

\iffalse
\subsubsection{Impact of Fine-Tuning Steps on Model Performance}
To see the effect of fine-tuning steps on the training of the Questioner, we present the BLEU and ROUGE scores in Figure \ref{fig:exp_steps}. In this figure, we show the performance of our Questioner without fine-tuning, as well as with $4000$, $28000$, and $40000$ steps.
An observation is that fine-tuning the model with more steps helps it to ask better follow-up questions, which is the goal of the Questioner. For example, in the non-fine-tuned model, after asking the third question, the model's performance remains the same or drops, indicating that it struggles to identify relevant follow-up questions. In contrast, the fine-tuned model with only $4000$ steps is able to ask effective follow-up questions, with each question up to question 6, helping the model to gather more user preferences.
 \fi

\subsubsection{Impact of Fine-Tuning Steps on Model Performance}
To see the effect of fine-tuning steps on the training of the Questioner, we present the BLEU and ROUGE scores in \cref{fig:exp_steps}. In this figure, we show the performance of our Questioner without fine-tuning, as well as with $4000$, $28000$, and $40000$ steps.
According to this figure, fine-tuning the model with more steps helps it to ask better follow-up questions, which is the goal of the Questioner. 

\iffalse
\textcolor{blue}{For example, in the non-fine-tuned model, after asking the third question, the model's performance remains the same or drops, indicating that it struggles to identify relevant follow-up questions. In contrast, the fine-tuned model with only $4000$ steps is able to ask effective follow-up questions, with each question up to question 6, helping the model to gather more user preferences.}\todo{[Ofer] This is not in the figure, did we intend to include another figure for this similar to \cref{fig:effect_questions}?}\fi

\subsubsection{Analyzing the Questions Asked by the Model}
Our aim in fine-tuning the Questioner to generate follow-up questions was to create a funnel format that aligns more closely with human-like conversation flow. Our training strategy, discussed above, %in \cref{sec:profile_corruption},
also involved generating funnel questions and removing information from the user profile based on these questions.

To determine whether the model is asking questions in a funnel format, we analyze the concept of each question (i.e., keywords in the JSON format of user profiles) within a conversation and calculate an expected value (or weighted rank) for each concept. This approach helps identify the order in which specific concepts are most likely to appear across conversations. The weighted rank (WR) for each concept is calculated as follows:
\begin{equation}
    WR = \sum_{i=1}^{T} i \times p(i)
\end{equation}
where $T$ is the maximum number of questions the model can ask, and  $p(.)$ is the probability that the concept appears in position $i$.

%\todoo{shouldn't this be also a function of the concept?}

The results of this experiment are shown in \cref{fig:exp_funnel_concepts}. 
According to this figure, the Questioner begins by asking broader concepts such as `Genre', `Film Era', and `Decade', gradually shifting to more specific questions like `Directors', `Visual Style', and `Tone', and eventually concluding with highly detailed questions, such as `Special Effects', `Humor', and `Atmosphere'. This progression suggests that the Questioner aims to start with broad concepts earlier in the conversation, then moves to more detailed questions as the conversation progresses, which is consistent with our data generation process.

\section{Related work}
Preference elicitation in conversational recommender systems plays a crucial role in quickly understanding user preferences and delivering tailored recommendations \citep{christakopoulou2016towards}. Recent research has focused on eliciting human preferences using language models (LMs) through various approaches. \citet{li2023eliciting} introduced Generative Active Task Elicitation (GATE), a framework where models interact with users through free-form language to infer intended behavior.
One interesting approach to enhancing language models involves teaching them to ask clarifying questions. This approach, known as STaR-GATE, aims to improve the performance of language models by enabling them to seek additional information when faced with ambiguity or uncertainty in a given task \citep{andukuri2024star}.
\cite{austin2024bayesian} introduced a framework for Bayesian optimization with LLM-based acquisition functions for natural language preference elicitation. The framework, demonstrated in the PEBOL (Preference Elicitation with Bayesian Optimization augmented LLMs) algorithm, utilizes Natural Language Inference (NLI) and Bayesian Optimization (BO) strategies, such as Thompson Sampling (TS) and Upper Confidence Bound (UCB), to steer LLM query generation.
\citep{piriyakulkij2023active} presents an algorithm for active preference inference using language models and probabilistic reasoning. By prompting instruction-tuned large language models with informative questions, the algorithm aims to enhance the ability of language models to quickly infer user preferences, transforming them into more interactive systems. \cite{shah2023using}  has explored intent classification through manual annotation and supervised learning, but these approaches often struggle to scale or adapt to the evolving nature of user interactions, especially in conversational AI settings.
\section{Conclusions}
We propose a novel model inspired by diffusion techniques to enhance large language models (LLMs) in generating funnel questions that effectively capture user preferences across various domains.
Our approach involves introducing noise into user profiles and training the model to denoise them by generating relevant questions.
Experimental results demonstrate significant improvements in the ability of LLMs to produce domain-specific, contextually appropriate follow-up questions.
We believe our model can advance personalized user interactions and open new avenues for adaptive learning in LLMs.

\bibliographystyle{ACM-Reference-Format}
\bibliography{sample-base}


\end{document}